\documentclass{article}

\usepackage[final, nonatbib]{neurips_2022}

\usepackage[utf8]{inputenc} 
\usepackage[T1]{fontenc}    
\usepackage{hyperref}       
\usepackage{url}            
\usepackage{booktabs}       
\usepackage{amsfonts}       
\usepackage{nicefrac}       
\usepackage{microtype}      
\usepackage{xcolor}         
\usepackage{graphicx}
\usepackage{amsmath}
\usepackage{amssymb}
\usepackage{booktabs}
\usepackage{float} 
\usepackage[font=small,labelfont=bf]{caption}

\title{FacEDiM: A Face Embedding Distribution Model for Few-Shot Biometric Authentication of Cattle}

\author{%
  Meshia C. Oveneke$^{1,2}$, Rucha Vaishampayan$^1$, Deogratias L. Nsadisa$^1$, Jenny A. Onya$^{1,2}$ \\
  Artificial Intelligence Research Lab\\
  $^1$Fit-For-Purpose Technologies SRL \& $^2$Neotex SARL\\
  \texttt{info@fitforpurpose.tech}, \texttt{info@neotex.ai} \\
}

\begin{document}

\maketitle

\begin{abstract}
  This work proposes to solve the problem of few-shot biometric authentication by computing the Mahalanobis distance between testing embeddings and a multivariate Gaussian distribution of training embeddings obtained using pre-trained CNNs.
  Experimental results show that models pre-trained on the ImageNet dataset significantly outperform models pre-trained on human faces. 
  With a VGG16 model, we obtain a FRR of $1.25\%$ for a FAR of $1.18\%$ on a dataset of 20 cattle identities. 
\end{abstract}

\section{Motivation} \label{sec:introduction}
According to recent statistics of the \textit{Food and Agriculture Organization} (FAO) of the \textit{United Nations} (UN), demand for milk and milk products in developing countries is significantly growing with rising incomes, population growth, urbanization and changes in diets \footnote{\url{https://www.fao.org/dairy-production-products/products/en/}}.
To face this continuous increase in demand, the proper authentication of cattle is essential, because it enables producers to keep comprehensive records for production, reproduction, health problems, and management practices. 
In the meantime, it is well known that traditional non-biometric methods for cattle authentication are not really satisfactory in providing reliability due to theft, fraud, and duplication.
Recently, biometrics based animal authentication using state-of-the-art \textit{Artificial Intelligence} (AI) techniques has therefore attracted more and more interest from the scientific community \cite{li2022mobilenetv2, xu2022cattlefacenet, kumar2016face}.
Since more than a decade, \textit{Deep Learning} (DL) has been successfully employed for vision-based face biometrics.
This is particularly due to the development of effective deep architectures and the release of quite consequent datasets.
However, one of the main limitations of DL is that it requires large scale annotated datasets to train efficient models. 
Collecting, annotating and managing such large datasets can be extremely time consuming and laborious, especially in areas where experts from the field are required, like in the animal science domain. 
The goal of this research is to investigate approaches to overcome such limitations by investigating ways to learn from few annotated data by transfering knowledge from DL models pre-trained on large computer vision datasets such as ImageNet \cite{krizhevsky2009learning}.

\section{Methodology and Experimental Results}
At the heart of our proposed methodology is the idea of posing the problem of biometric authentication as an anomaly detection problem in the biometric template space. Inspired by recent advances in anomaly detection \cite{defard2021padim}, we propose to model a per-identity multi-variate Gaussian distribution of face embeddings and perform biometric authentication by measuring the deviation from this distribution, hence \textit{Face Embedding Distribution Model} (FacEDiM).
To alleviate the data scarcity problem, we automatically generate samples by augmenting the few shots using a method similar to \cite{dosovitskiy2015discriminative}, i.e. applying image transformation such as scaling, rotation, translation, color variation, and contrast variation.
For extracting the face embeddings from a few shots, we make use of pre-trained \textit{convolutional neural networks} (CNN)s.
Let's define the $d$-dimensional face embedding function $\mathbf{f}: \mathbb{R}^{h \times w \times c} \to \mathbb{R}^d$ as follows: $\mathbf{f}(\mathbf{x}) \triangleq \mathbf{f}_E \circ \ldots \circ \mathbf{f}_1(\mathbf{x})$, where $\mathbf{x}$ denotes a 2D input image with dimensions $h \times w $ and $c$ color channels, and $\mathbf{f}_E$ denotes the embedding layer of the CNN, usually corresponding to a fully-connected layer.
We then define a family of transformations $\{\mathbf{T}_{\boldsymbol{\alpha}} | \boldsymbol{\alpha} \in \mathcal{A}\}$, parameterized by a random vector $\boldsymbol{\alpha} \in \mathcal{A}$, where $\mathcal{A}$ is the set of all possible parameter vectors.
We model a multivariate Gaussian distribution $\mathcal{N}(\boldsymbol{\mu}, \boldsymbol{\Sigma})$ to get a probabilistic representation of the reference face template.
The mean $\boldsymbol{\mu} \in \mathbb{R}^d$ is estimated as follows:
\begin{equation} \label{eq:mean}
		\boldsymbol{\mu} = \mathbb{E}_{p(\mathbf{x})}\mathbb{E}_{p(\boldsymbol{\alpha})}[\mathbf{f}(\mathbf{T}_{\boldsymbol{\alpha}} (\mathbf{x}))] \approx \frac{1}{MN}\sum_{i=1}^{M}\sum_{j=1}^{N} \mathbf{f}(\mathbf{T}_{\boldsymbol{\alpha}_j} (\mathbf{x}_i))
\end{equation}
The sample covariance $\boldsymbol{\Sigma} \in \mathbb{R}^{d \times d}$ is estimated as follows:
\begin{equation} \label{eq:covariance}
	\begin{split}
		\boldsymbol{\Sigma} & = \mathbb{E}_{p(\mathbf{x})}\mathbb{E}_{p(\boldsymbol{\alpha})}[\left(\mathbf{f}(\mathbf{T}_{\boldsymbol{\alpha}} (\mathbf{x})) - \boldsymbol{\mu}\right) \left(\mathbf{f}(\mathbf{T}_{\boldsymbol{\alpha}} (\mathbf{x})) - \boldsymbol{\mu}\right)^T] \\
		& \approx \frac{1}{MN-1}\sum_{i=1}^{M}\sum_{j=1}^{N} \left(\mathbf{f}(\mathbf{T}_{\boldsymbol{\alpha}_j} (\mathbf{x}_i)) - \boldsymbol{\mu}\right) \left(\mathbf{f}(\mathbf{T}_{\boldsymbol{\alpha}_j} (\mathbf{x}_i)) - \boldsymbol{\mu}\right)^T  + \epsilon\mathbf{I}_d
	\end{split}
\end{equation}
where the regularization term $\epsilon\mathbf{I}_d$ makes the sample covariance matrix $\boldsymbol{\Sigma}$ full rank and invertible.
Having this probabilistic representation, the distance between a test image $\mathbf{x}' \in \mathbb{R}^{h \times w \times c}$ and the learned face embedding distribution $\mathcal{N}(\boldsymbol{\mu}, \boldsymbol{\Sigma})$ is computed using the Mahalanobis distance \cite{mahalanobis1936generalized}:
\begin{equation} \label{eq:distance}
	d_{\boldsymbol{\Sigma}}(\mathbf{x}',\boldsymbol{\mu}) = \sqrt{\left(\mathbf{f}(\mathbf{x}') - \boldsymbol{\mu}\right) \boldsymbol{\Sigma}^{-1} \left(\mathbf{f}(\mathbf{x}') - \boldsymbol{\mu}\right)^T}
\end{equation}

\includegraphics[width=\textwidth]{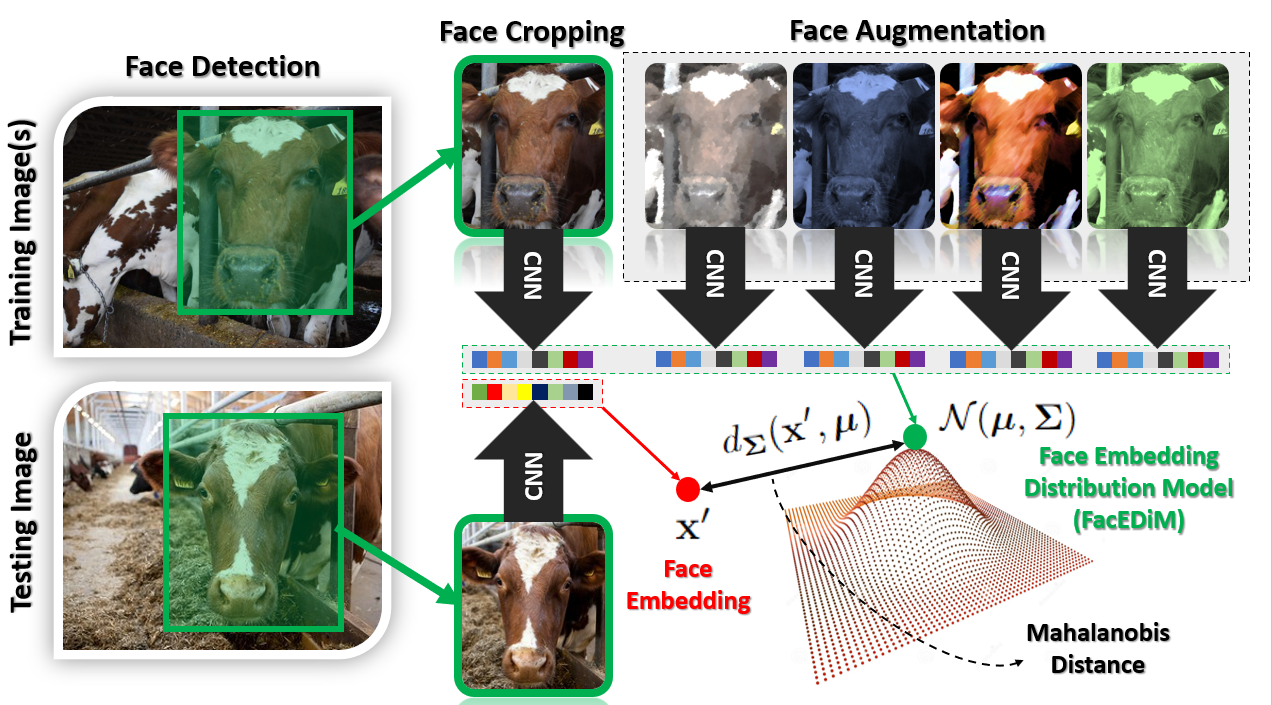}
\captionof{figure}{Proposed FacEDiM framework.}
\label{fig:facedim}

This relatively simple method (\ref{eq:distance}) allows us detecting out-of-distribution biometric templates given a pre-defined distance threshold.
The more augmentations $N$ and initial shots $M$, the more robust the distribution mean (\ref{eq:mean}) and covariance (\ref{eq:covariance}).
Figure \ref{fig:facedim} gives a overview of our proposed FacEDiM framework. 
The input images $\mathbf{x}$ are first processed with a publicly available cattle face detector \footnote{ \url{https://rapidapi.com/FitForPurposeTechnologies/api/halisi-animal-face-detection/}}.
The resulting bounding boxes are used for cropping the face region and extracting the face embeddings $\mathbf{f}(\mathbf{x})$ using a pre-trained CNN.
A per-identity distribution is then modeled using the few training shots and biometric authentication is performed using the Mahalanobis distance on test embeddings.

\begin{table}[h!]
	\centering
	\begin{tabular}{|l|c|c|c|} 
		\hline
		\textbf{Backbone CNN} & \textbf{FAR} & \textbf{FRR}& \textbf{EER} \\ 
		\hline\hline
		\textbf{VGG16} \cite{simonyan2014very} & \textbf{0.0118} & \textbf{0.0125} & \textbf{0.0121}\\  
		ResNet50 \cite{he2016deep} & 0.0063 & 0.0285 & 0.0174\\
		MobileNetV2 \cite{sandler2018mobilenetv2} & 0.0033 & 0.0578 & 0.0305\\
		DenseNet121 \cite{huang2017densely} & 0.0154 & 0.0200 & 0.0177\\
		EfficientNetB0 \cite{tan2019efficientnet} & 0.0027 & 0.0437 & 0.0232\\
		\hline
		VGG-Face \cite{tan2019efficientnet} & 0.0445 & 0.0600 & 0.0523\\
		FaceNet \cite{schroff2015facenet} & 0.2619 & 0.0802 & 0.1711 \\
		FaceNet512 \cite{tan2019efficientnet} & 0.2006 & 0.1350 & 0.1678\\
		DeepFace \cite{taigman2014deepface} & 0.0006 & 0.1410 & 0.0708 \\
		DeepID \cite{sun2014deep} & 0.1355 & 0.0490  & 0.0923 \\
		\hline
	\end{tabular}
	\captionof{table}{Experimental results.}
	\label{tbl:results}
\end{table}

To empirically evaluate our proposed framework, we've used five backbone CNNs pre-trained on the ImageNet dataset: VGG16 \cite{simonyan2014very}, ResNet50 \cite{he2016deep}, MobileNetV2 \cite{sandler2018mobilenetv2}, DenseNet121 \cite{huang2017densely} and EfficientNetB0 \cite{tan2019efficientnet}.
To assess the transferability from human to cattle biometrics, we've used the following backbone CNNs pre-trained on human faces: VGG-Face \cite{tan2019efficientnet}, FaceNet \cite{schroff2015facenet}, FaceNet512 \cite{tan2019efficientnet}, DeepFace \cite{taigman2014deepface} and DeepID \cite{sun2014deep}.
Table \ref{tbl:results} summarizes the results we've obtained with an in-house dataset containing 20 cattle identities, with number of shots per identity $M = 10$. 
We then augmented the 10 shots by a factor of $N = 100$ and divided the resulting dataset into 800 shots for training and 200 shots for testing.
For each backbone CNN, we then calculated the Mahalanobis distances for each of the 200 testing shots against the 10 distributions obtained using the 800 training shots.
An optimal distance threshold is obtained by observing the best \textit{Equal Error Rate} (EER), i.e. trade-off between \textit{False Rejection Rate} (FRR) and \textit{False Acceptance Rate} (FAR). 
Surprisingly, with a VGG16 model pre-trained on the ImageNet dataset we obtain the best results, i.e. FRR of $1.25\%$ for a FAR of $1.18\%$. 
These very encouraging results prove that we can obtain state-of-the-art cattle authentication results by overcoming the limitations caused by time consuming and laborious cattle data collection and annotation. 
As a future work, we will extend this framework to other animals.

\section*{Acknowledgements}
This work is funded by Fit-For-Purpose Technologies SRL, \url{www.fitforpurpose.tech/halisi}.

\small
\bibliographystyle{plain}
\bibliography{references}

\end{document}